\documentclass[10pt,twocolumn,letterpaper]{article}

\usepackage[pagenumbers]{cvpr} 

\usepackage{graphicx}
\usepackage{amsmath}
\usepackage{amssymb}
\usepackage{booktabs}
\usepackage[accsupp]{axessibility}  

\newcommand\blfootnote[1]{%
  \begingroup
  \renewcommand\thefootnote{}\footnote{#1}%
  \addtocounter{footnote}{-1}%
  \endgroup
}

\usepackage[pagebackref,breaklinks,colorlinks]{hyperref}

\usepackage[capitalize]{cleveref}
\crefname{section}{Sec.}{Secs.}
\Crefname{section}{Section}{Sections}
\Crefname{table}{Table}{Tables}
\crefname{table}{Tab.}{Tabs.}

\def\cvprPaperID{29} 
\def\confName{CVPR}
\def\confYear{2023}

\begin{document}

\title{Disentangling Neuron Representations with Concept Vectors}  

\author{Laura O'Mahony\thanks{corresponding author lauraa.omahony@ul.ie } $^{1,2}$\\ 
{\tt\small lauraa.omahony@ul.ie}
\and
Vincent Andrearczyk$^1$\\
{\tt\small vincent.andrearczyk@hevs.ch}
\and
Henning M\"uller$^{1,3}$\\
{\tt\small henning.muller@hevs.ch}
\and
Mara Graziani$^1$\\
{\tt\small mara.graziani@hevs.ch}\\
$^1$ Haute école spécialisée de Suisse occidentale, Hes-so Valais, Sierre, Switzerland\\
$^2$ University of Limerick, Limerick, Ireland\\
$^3$ The Sense Research and Innovation Center, Sion, Lausanne, Switzerland\\
}

\maketitle

\begin{abstract}
   
   Mechanistic interpretability aims to understand how models store representations by breaking down neural networks into interpretable units. However, the occurrence of polysemantic neurons, or neurons that respond to multiple unrelated features, makes interpreting individual neurons challenging. This has led to the search for meaningful vectors, known as concept vectors, in activation space instead of individual neurons. 
   The main contribution of this paper is a method to disentangle polysemantic neurons into concept vectors encapsulating distinct features. Our method can search for fine-grained concepts according to the user's desired level of concept separation. The analysis shows that polysemantic neurons can be disentangled into directions consisting of linear combinations of neurons. Our evaluations show that the concept vectors found encode coherent, human-understandable features. 
\end{abstract}

\section{Introduction}
\label{sec:intro}

\blfootnote{Source code available at \url{https://github.com/lomahony/sw-interpretability}}
\blfootnote{Proceedings of the IEEE / CVF Conference on \textit{Computer Vision and Pattern Recognition Workshops} 2023, Vancouver, Canada. Copyright 2023 by the author(s).}

Mechanistic interpretability is a fast-emerging research topic that aims at deciphering the internal representations held by a model by reverse engineering into understandable computer programs~\cite{olah2017feature, olah2020zoom, agarap2018deep}. 
Previous work in this field breaks down convolutional neural networks (CNNs) into the features learned by the fundamental units of a layer, which are considered as 
directions of a geometric space. 
Many previous works consider neurons as these units~\cite{olah2017feature, olah2020zoom}. 
Breaking down the model into such interpretable units allows us to better understand how models store representations in vision tasks~\cite{olah2017feature, olah2020zoom, bau2017network, bau2020understanding} and language models~\cite{elhage2021mathematical}. This could even allow us to predict and edit model behaviour such as work by Bau \etal that removes units that are important to a class~\cite{bau2020understanding} and has also been studied for other architectures such as GANs and GPT models~\cite{bau2020rewriting, meng2022locating, meng2022mass}. 

A frequent issue is the occurrence of \textit{polysemantic neurons}, namely neurons that respond to several unrelated features, or concepts~\cite{olah2017feature, olah2020zoom, nguyen2016multifaceted}. 
They can be found by looking at the maximally activating dataset examples and finding they consist of multiple groups that are conceptually very different~\cite{olah2020zoom, black2022interpreting}. 
This makes the interpretation of individual neurons challenging since they cannot be mapped to unique features. This is exemplified in Olah \etal ~\cite{olah2017feature, olah2020zoom} by a neuron equally likely to respond to car shields and cat paws at the same time, and with the same intensity.
There is evidence that the training of models pushes networks to represent many features within individual neurons~\cite{scherlis2022polysemanticity, elhage2022toy}. Models have a limited number of neurons meaning a discrete neuron is often not possible for all features. 
This is related to the idea of \textit{superposition}, which refers to when neural networks represent more features than they have neurons~\cite{elhage2022toy}. 
These empirical observations in existing research indicate that neurons are not always the right fundamental unit encapsulating an individual feature represented by a model. 
If we define activation space as all possible combinations of neuron activations, we can widen our lens to look for meaningful vectors in activation space instead of single neuron basis vectors. 

There is evidence that suggests monosemantic regions in activation space exist~\cite{olah2017feature, bau2017network, elhage2022toy, black2022interpreting, jermyn2022engineering}, 
but they are not always made obvious by studying individual neurons~\cite{black2022interpreting, szegedy2013intriguing}. 
A key issue resulting from this observation is the question of 
how directions in activation space representing distinct features 
can be found~\cite{olah2017feature}.

\begin{figure*}[ht!]
  \centering
    \includegraphics[width=1\textwidth]{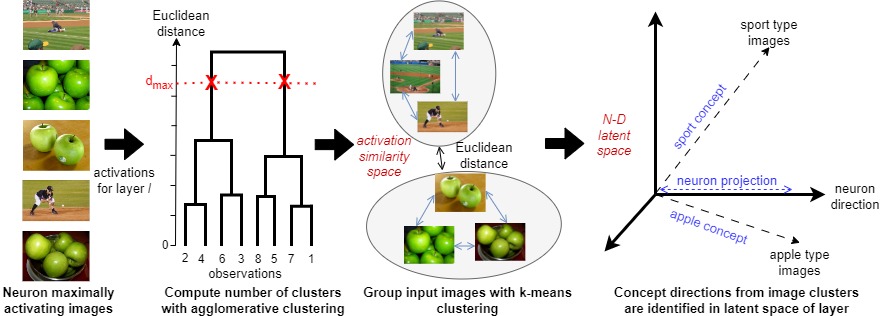}
  \caption{Step 1. A set of images that maximally activate a neuron in a model layer is taken. Step 2. The Euclidean distance between images in activation space is used as the similarity space on which the clustering is performed. This returns the appropriate number of clusters for a given distance threshold. Step 3. K-means clustering computes the cluster membership. Step 4. From the images in each cluster, a concept vector is calculated, which points toward the non-neuron aligned direction in activation space. }
  \label{fig:method}
\end{figure*}

Previous work has shown the existence of high-level human interpretable concepts such as textures, shapes and parts of objects present as directions in activation space. Some early work by Alain \etal~\cite{alain2016understanding} took the features of the layers in a model and fit a linear classifier to each layer to predict the class labels. The work by Kim \etal~\cite{kim2018interpretability} on \textit{Concept Activation Vectors (CAVs)} defines a concept as a vector in the direction of the activation values of a set of examples of that concept. The authors find a concept by training a linear classifier to distinguish between examples of that concept and random counterexamples. The concept vector is then taken as the vector orthogonal to the boundary. A limitation of this method is that it requires a handcrafted set of examples of a concept to find the concept direction in latent space. A small number of unsupervised methods used to find concepts have been developed~\cite{raghu2017svcca, graziani2023concept, mcgrath2022acquisition}, this research direction is known as \textit{concept discovery}. 

Concept discovery involves the search for unit vectors in the latent space of a model that encode learned representations of high-level concepts. However, none of the existing methods seek to disentangle polysemantic representations. 
The concept vectors are linear combinations of units, and as such, they are likely to inherit polysemanticity from polysemantic neurons~\cite{graziani2023concept}. 
Furthermore, none of these methods incorporate the notion of a privileged basis proposed by Elhage \etal~\cite{elhage2022toy}. A \textit{privileged basis} is where some representations are encouraged to align with basis directions, meaning directions in space corresponding to individual neurons. 
Even though neuron directions are usually meaningful candidates for representing a feature~\cite{olah2017feature, olah2020zoom, bau2017network, elhage2022toy}, they likely do not show the whole story due to the countervailing force of superposition~\cite{elhage2022toy}. The main contribution of this paper is a method to find and disentangle monosemantic directions starting from polysemantic neurons. Moreover, our method can search for concepts that are fine-grained according to the user’s desired level of concept separation. Our analysis shows that polysemantic neurons can be disentangled into directions 
consisting of linear combinations of neurons. 

\section{Methods}
\label{sec:methods}

We consider a CNN predicting a classification output ($p$-dimensional output vector) from an input image. 
We note that the method can be generalised to other models, but use a CNN for our analysis. 
We assume the model was already trained, and that we have access to the intermediate representations of an arbitrary layer inside the model. 
\cref{fig:method} summarises the steps discussed in more detail below.

We take a given intermediate layer $l$. 
We calculate the embeddings for the entire dataset and apply global average pooling to aggregate the spatial information of the convolutional feature maps. 
We select a neuron $n$,  
and apply the following steps iteratively. 
In step 1, we take these activations 
$ \{\phi^l(x_i)\}_{i=1}^N$ where $\phi^l(x_i) \in \mathbb{R}^{d} $, and for the neuron $n$, we 
take the top $N$ activating images $\{x_i\}_{i=1}^N $.

The second step involves measuring the similarity of the pooled activations (of the top activating images) at the intermediate layer $l$. We use the Euclidean distance as a distance metric which has been shown by previous work to be highly predictive of perceptual similarity~\cite{zhang2018unreasonable}. We then 
apply a clustering technique to group these measurements of similarity into sets of close examples. For this, we use agglomerative clustering, a bottom-up type of hierarchical clustering~\cite{murtagh2012algorithms}, since it does not require us to pre-specify the number of clusters to be generated, as is required by the k-means approach. 
With clustering settings described in~\cref{apn:extension}, we apply agglomerative clustering with a distance threshold $d_{max}$ that specifies the maximum linkage threshold at which clusters will be merged. 
Its result can be visualised in a dendrogram, or tree-based representation of elements, as is depicted in \cref{fig:method}, step 2.
The distance threshold is a hyperparameter that is tuned to an appropriate range for the model layer. It can be tweaked to fine-grain concepts into big or small buckets. 
Please refer to Appendix~\ref{apn:extension} for a demonstration of this.

The third step takes the resulting number of clusters $C$ obtained from step 2 and performs k-means clustering on the same measurements of similarity. The benefit of using k-means clustering over agglomerative clustering alone is that the k-means centroids allow us to easily remove outliers from each cluster that have low similarity to the rest of the cluster as employed in~\cite{ghorbani2019towards}. 

The final step finds the directions of the $\hat{C}$
\footnote{$\hat{C}$ may be different from $C$ since clusters with less than 5 samples are removed.} 
concept vectors corresponding to the $\hat{C}$ clusters, $\{\hat{c_j}\}_{j=1}^{\hat{C}}$, by taking the mean of the remaining embeddings for each cluster, giving us a set of vectors $ \{ avg( \{\phi^l(x_i)\}_{i \in \hat{c_j}} ) \}_{j=1}^{\hat{C}} $ which are then normalised to give us disentangled concept vectors $ \{ v_j\}_{j=1}^{\hat{C}} $, where $v_j \in \mathbb{R}^{d} $.

\section{Results}
\label{sec:results}

We consider Inception V3 (IV3)~\cite{szegedy2016rethinking} for our experiments since it is a de-facto standard convolutional neural network. Moreover, interpretability research has already given multiple insights for this model~\cite{kim2018interpretability, ghorbani2019towards, graziani2023concept}, and pretrained weights on the ImageNet ILSVRC2012~\cite{russakovsky2015imagenet} dataset are available online. 
As this exploratory study only aims at a proof of concept, we focused on an undersampled version of ImageNet, retaining 130 random images for each class. This kept computation accessible to our infrastructure, feasible and light. Our results can easily be scaled to the entire dataset and larger dataset sizes. Where not stated, we consider the concatenation layer \textit{Mixed 7b}, a convolutional layer with 2048 feature maps ($d$ = 2048) near the end of the IV3 model. We pick this layer as we expect it to encode complex concepts~\cite{zeiler2014visualizing, olah2017feature}. A similar analysis can be done on other layers and architectures.

\begin{figure}
  \centering
  \begin{subfigure}{0.48\linewidth}
    \includegraphics[width=1\textwidth]{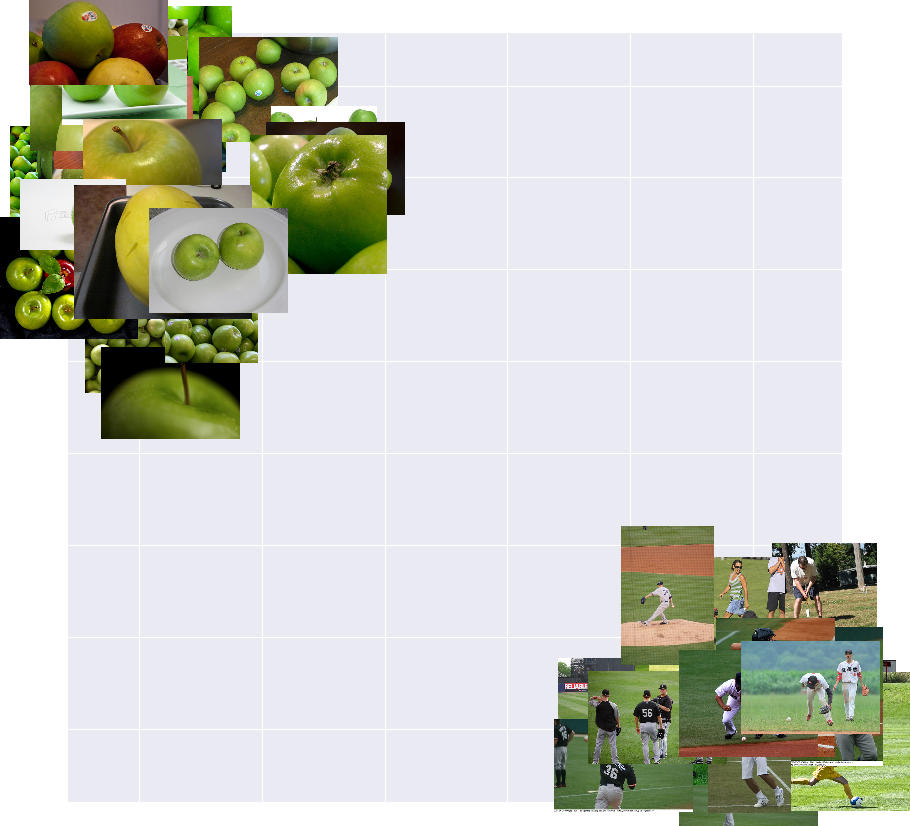}
    \caption{}
    \label{fig:kmeans_clean_35}
  \end{subfigure}
  \hfill
  \begin{subfigure}{0.48\linewidth}
  \centering
    \includegraphics[width=1\textwidth]{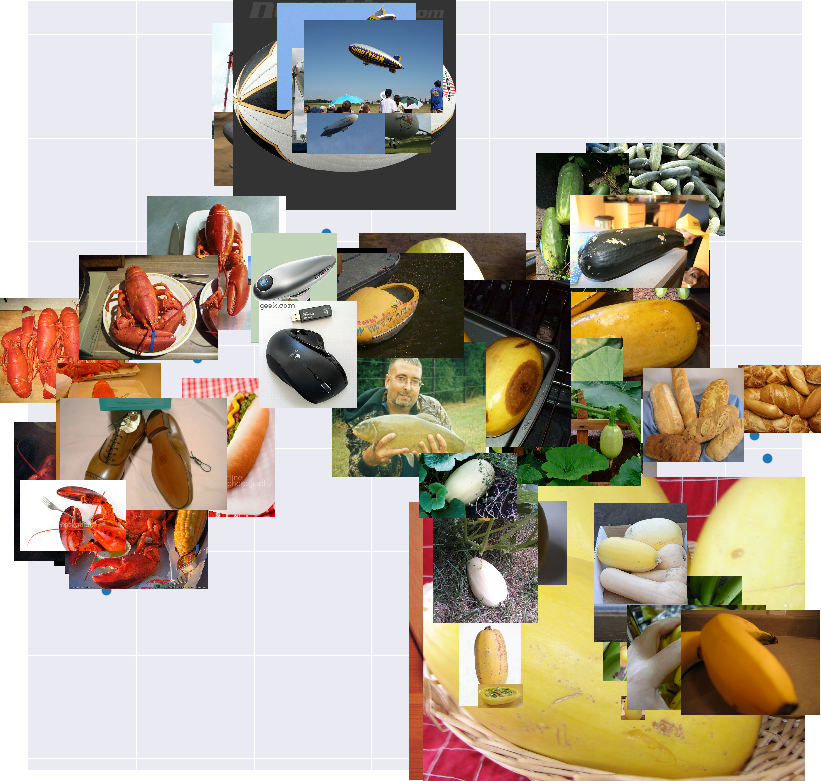}
    \caption{}
    \label{fig:kmeans_clean_16}
  \end{subfigure}
  \begin{subfigure}{0.5\linewidth}
    \includegraphics[width=1\textwidth]{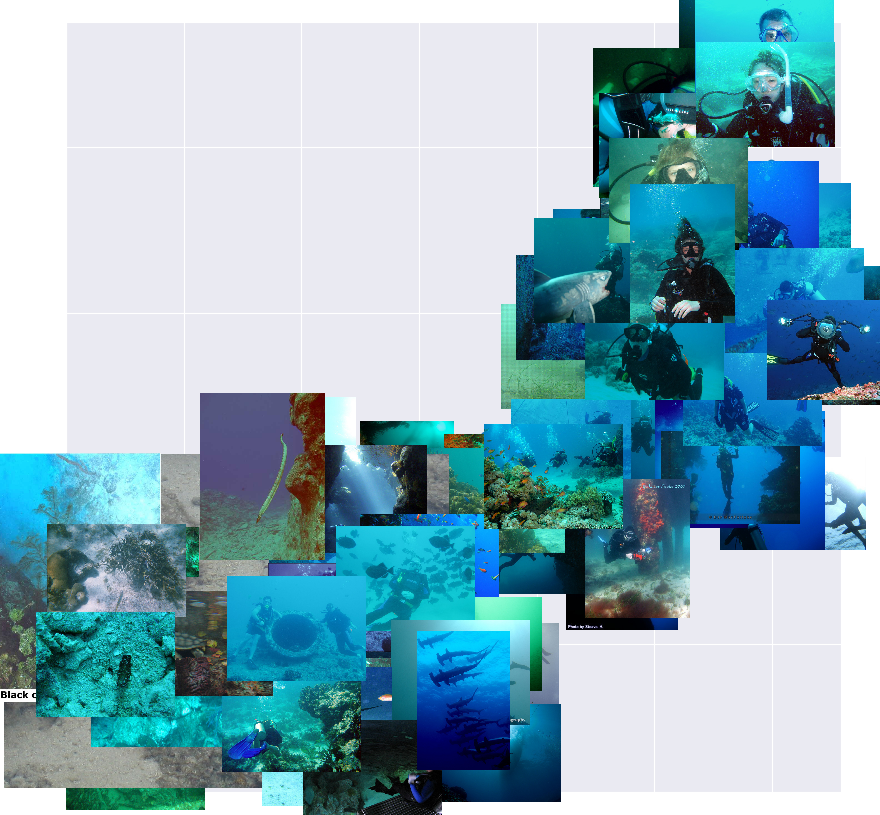}
  \caption{}
  \label{fig:kmeans_clean_1}
  \end{subfigure}
  \caption{UMAP of the maximally activating images kept after k-means clusters and outlier removal in latent space: (a) separate clusters for polysemantic neuron 35 (b) a single cluster for monosemantic neuron 16.
  }
  \label{fig:kmeans_clean}
\end{figure}

We demonstrate the results of the method described in \cref{sec:methods} on a number of both polysemantic and monosemantic neurons. We took $N$ = 100 top activating dataset examples and set the distance threshold parameter $d_{max}$ = 15. 
We select neuron 35 as an example of a polysemantic neuron, as it activates highly for images of apples, sports, and also three images are dominated by a net-like pattern. This results in 3 clusters. When the k-means clustering step was applied (i.e. Step 3 in Figure~\ref{fig:method}) and outliers were removed, the cluster containing the net-like images was removed as there were $<$ 5 images in this cluster. \cref{fig:kmeans_clean_35} shows the embeddings of the remaining images plotted using UMAP~\cite{mcinnes2018umap} dimensionality reduction. The plot shows how UMAP separates embeddings for images of apples and sports. Note that UMAP is used only for demonstration of our results to depict the clusters found using k-means since we found it accurately reflects the cluster membership result. We select neuron 16 as an example of a monosemantic neuron that activates for many categories of elliptical shapes as depicted in \cref{fig:kmeans_clean_16}. The same procedure applied to this neuron yields only one cluster, yielding one monosemantic concept vector which has a much higher similarity than the neuron direction to the original images. 
The case of neuron 1 is interesting and demonstrates the application of our method to fine-grain concepts. This neuron activates highly for underwater images as shown in \cref{fig:kmeans_clean_1}. Further inspection shows how it activates highly for both general underwater images and, images of scuba divers. Applying our method yields two clusters, one for general underwater scenes such as coral, and another for scuba divers, meaning two concept vectors can be found for this neuron. However, at a higher distance threshold, the clusters of images are merged by the algorithm, and one concept is found. 
Here the features are not entirely conceptually different. However, these two related directions can be differentiated, and concept discovery can separate them. We believe this hints at how it may be helpful to view concepts as varying continuously in the latent space instead of being encoded discretely by neurons. We suspect this phenomenon is related to the notion of `feature facets'~\cite{nguyen2016multifaceted}. 
The same analysis was scaled to multiple neurons in the same layer. We provide 
additional examples and a depiction of the dendrogram for neuron 1 to see the result of 
altering $d_{max}$ in the Appendix~\ref{apn:extension}, \cref{fig:kmeans_clean_apn,fig:dendrogram_1}. 
We note that we found that the majority of the neurons we analysed in layer \textit{Mixed 7b} were found to show some amount of polysemanticity. A possible explanation for this is that 
the number of features may be very high for a later layer in the model as it encodes complex concepts.

\begin{figure}
  \centering
    \includegraphics[width=0.46\textwidth]{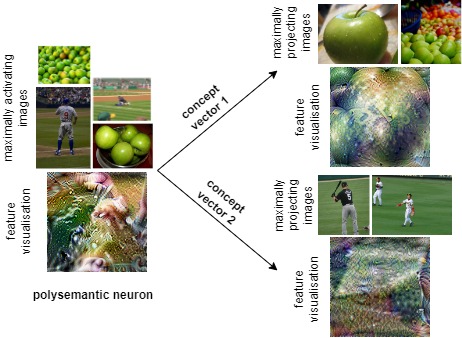}
  \caption{Disentanglement of the representations for neuron 35. The maximally activating inputs and feature visualisation are shown for the polysemantic neuron (left) and the disentangled concept directions (right).}
  \label{fig:feature_visualisation}
\end{figure}

We performed a qualitative and quantitative assessment of the identified concepts. The semanticity of concept vectors was evaluated, first by finding the dataset examples with the largest projections along the vectors, analogous to viewing the maximally activating dataset examples for individual neurons. \cref{fig:feature_visualisation} demonstrates how the two concept vectors found for polysemantic neuron 35 were confirmed to be monosemantic regions in the latent space, cleanly activated by the originally entangled concepts. A further qualitative assessment involved applying the technique of feature visualisation~\cite{olah2017feature} using the Lucent~\cite{lucent} library. \cref{fig:feature_visualisation} also shows the result of applying this technique on polysemantic neuron 35, and on the concept vectors found with our method. The polysemantic neuron fails to give a human interpretable representation, whereas the disentangled directions closer resemble the distinct categories of images which excite this neuron. For instance, the concept pointing towards representations of apples is maximally activated by round shapes with a stem cavity that is typical of apples, whereas the second concept seems to be maximally activated by large green squares such as football, soccer or baseball pitches.

As has been done in concept discovery works~\cite{ghorbani2019towards, graziani2023concept}, we evaluated our results with human experiments to evaluate the coherency and understandability of concepts. To avoid any cherry-picking of results, concepts from the first 8 neurons were used for all questions on the form, and a random number generator was used to select images from each concept's maximally projecting images\footnote{The evaluation form can be accessed at~\url{https://forms.gle/62H6iUiXHdLYs9nr9}}. The first four questions evaluated the concepts' coherency by asking participants to identify an intruder out of four other images that maximally activate another concept vector found from the same neuron or concept discovery starting point. 
The ($n=$ 8, including 3 domain experts) participants selected the intruder image with an overall 100\% accuracy showing the images have a coherent theme. The other six questions were designed to evaluate the understandability of concepts. Participants were asked to label two concepts and assess 
whether they agree with a given label for four sets of images. Agreement with the given labels was observed 97\% of the time. Further details are provided in Appendix~\ref{apn:user_evaluation} for the only label change suggestion (from a domain expert) that occurred. 
This confirms that the images have a consistent semantic meaning across multiple individuals.

\begin{figure}
  \centering
    \includegraphics[width=0.48\textwidth]{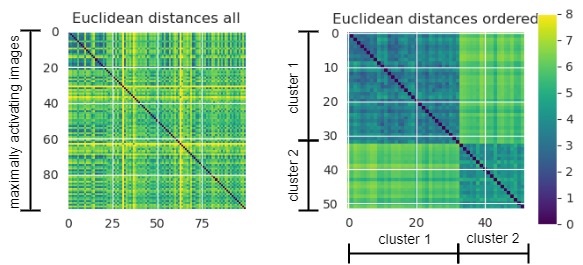}
  \caption{
  Euclidean distance between activation pairs, for the top 100 maximally activating images as in step 1 (on the left) and for the 52 remaining images after step 3, ordered by cluster membership (on the right). 
  }
  \label{fig:Euclidean_distances}
\end{figure}

\begin{figure}
  \centering
  \includegraphics[width=.46\textwidth]{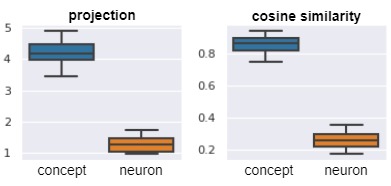}
  \caption{The left plot shows the projection of the elements of a cluster along the corresponding concept vector and the projection along the neuron direction for neuron 35. The right plot shows the cosine similarities between elements of a cluster with the concept vector and the neuron direction.}
  \label{fig:projection_cosine_35_cl1}
\end{figure}

A number of quantitative metrics were used to analyse the images making up the clusters for computing the concept vectors and also the maximally projecting images along concepts. A natural starting point was to check the distribution of Euclidean distances between maximally activating images as is shown in \cref{fig:Euclidean_distances} (a). \cref{fig:Euclidean_distances} (b) shows how the inter-cluster distance (distance between images of apples and sport-type images) is considerably higher than the intra-cluster distance. 
\cref{apn:concept_vectors} and particularly ~\cref{fig:concept_vector_35} illustrate the components of the 2048 concept vectors for these two concepts, which differ considerably across other dimensions. 
When the maximally projecting images along the concept vectors were calculated, our analysis confirmed that the projections and cosine similarities of their corresponding activations with the concept vector are remarkably higher than that with the neuron direction as exhibited in \cref{fig:projection_cosine_35_cl1}. 
As shown in additional examples in~\cref{apn:extension}, our results are consistent for other monosemantic and polysemantic neurons. 

\section{Conclusions and Future Work}
\label{sec:conclusion}

Finding meaningful directions in activation space that are pointing to unique patterns, or concepts is a non-trivial problem encountered in our journey of understanding neural networks. 
Our results suggest that exploring directions, instead of neurons may lead us toward finding coherent fundamental units. 
We believe this work helps move toward bridging the gap between understanding the fundamental units of models as is an important goal of mechanistic interpretability, and concept discovery. 
We evaluated the coherency and understandability of the raw images whose embeddings have the maximum projections along a concept vector. 
We found that the latent space representations have much 
higher similarities with the concept vectors discovered than with the neuron directions. 
This work goes in the direction of building interpretability in a human-controlled way, as is important for the field of AI safety, and for applications of image models such as medical lesion analysis. 
We note a limitation of this work is its reliance on the data used to generate clusters. Furthermore, all experiments were performed on image data as image data is easier to visualise than other data forms. 
Generalising the method to other data types such as language and tabular data is a direction we wish to pursue in future work, as is looking at other starting candidates for concepts besides neurons. 

\subsubsection{Acknowledgements} 
This work was supported by AI4Media of the European Union’s Horizon 2020 research and innovation
program under grant agreement No 951911. 
LOM performed this work during an internship facilitated by the AI4Media Junior Fellows Exchange Program. 
LOM also acknowledges the support of Science Foundation Ireland under Grant Number 18/CRT/6049. For the purpose of Open Access, the authors have applied a CC BY public copyright licence to any Author Accepted Manuscript version arising from this submission.


\newpage
{\small
\bibliographystyle{ieee_fullname}
\bibliography{Bibliography}

\newpage
\appendix

\renewcommand\thefigure{\thesection.\arabic{figure}}   
\section{Appendix}
\subsection{Further details of clustering and demonstrations of clustering results}
\label{apn:extension}

Ward linkage was used as the criterion to determine the merges between clusters in agglomerative clustering. 
This criterion minimises the variance within clusters by merging the pair of clusters with the minimum between-cluster sum of squared differences. 
 \cref{fig:dendrogram_35} depicts how many clusters agglomerative clustering selects for neuron 35. The height measures the distance between clusters, so the heights show the distances at which merges occur. The number of clusters $C$ can be inferred by placing a horizontal line at the desired distance or similarity threshold. 
 In the case of this neuron, the dendrogram shows there will be 2 clusters for around 16-28 and more for $<$ 16. At $d_{max}$ = 15 we get 3 clusters, if we pick a higher threshold, we would obtain 2 clusters. In step 3, Ward linkage is again used, so the k-means clustering works similarly to agglomerative clustering.

\begin{figure}[h]
  \centering
    \includegraphics[width=0.5\textwidth]{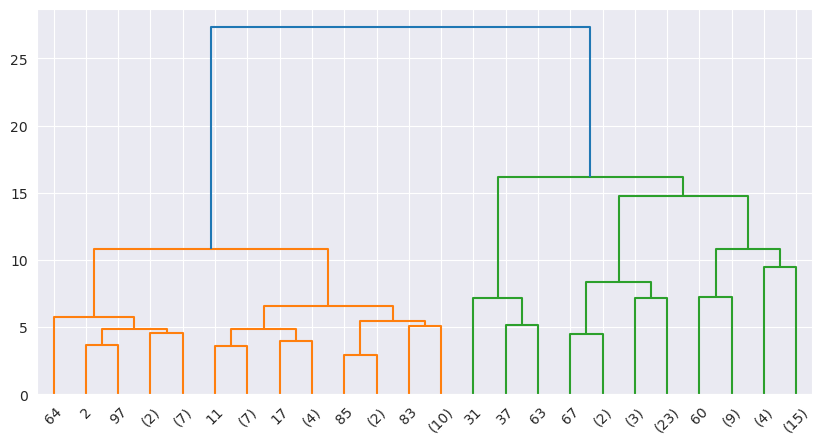}
  \caption{Neuron 35, agglomerative clustering dendrogram. }
  \label{fig:dendrogram_35}
\end{figure}

\cref{fig:kmeans_clean_5} shows the clustering result for polysemantic neuron 5. This neuron activates highest for images of peacocks and vans. Applying our method yields two disentangled vectors. 
\cref{fig:kmeans_clean_13} shows the clustering result for monosemantic neuron 13. This neuron activates the most for images with dark backgrounds. Applying our method yields one concept vector. 
An additional example of a polysemantic neuron is shown in \cref{fig:kmeans_clean_27}. Neuron 27 activates highest for images of toilet paper rolls and the faces of a specific dog breed.

\begin{figure}
  \centering
  \begin{subfigure}{0.5\linewidth}
    \includegraphics[width=1\textwidth]{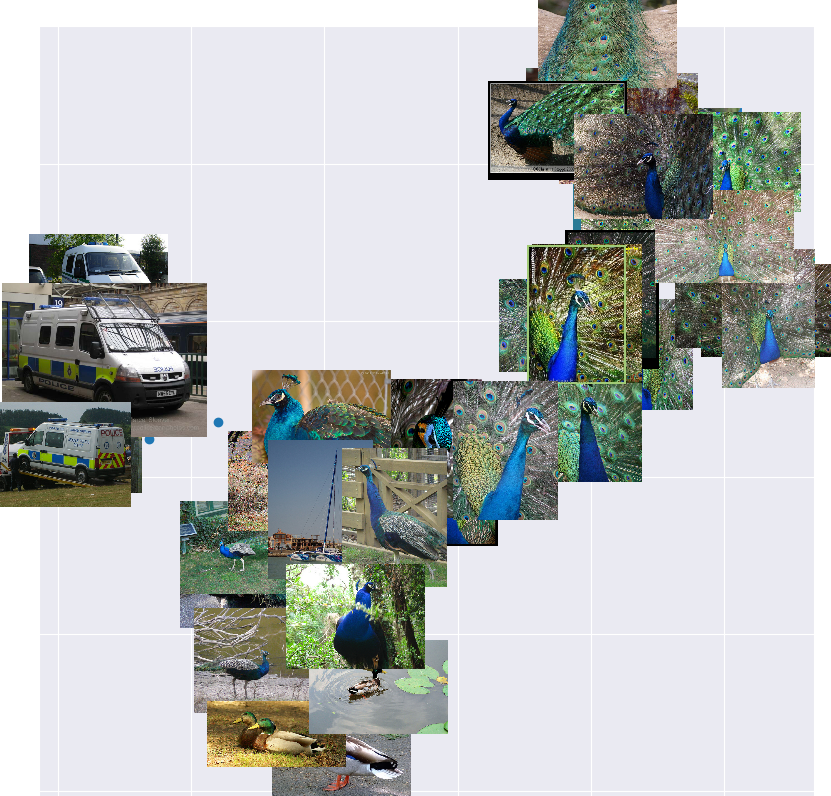}
  \caption{}
  \label{fig:kmeans_clean_5}
  \end{subfigure}
  \begin{subfigure}{0.5\linewidth}
    \includegraphics[width=1\textwidth]{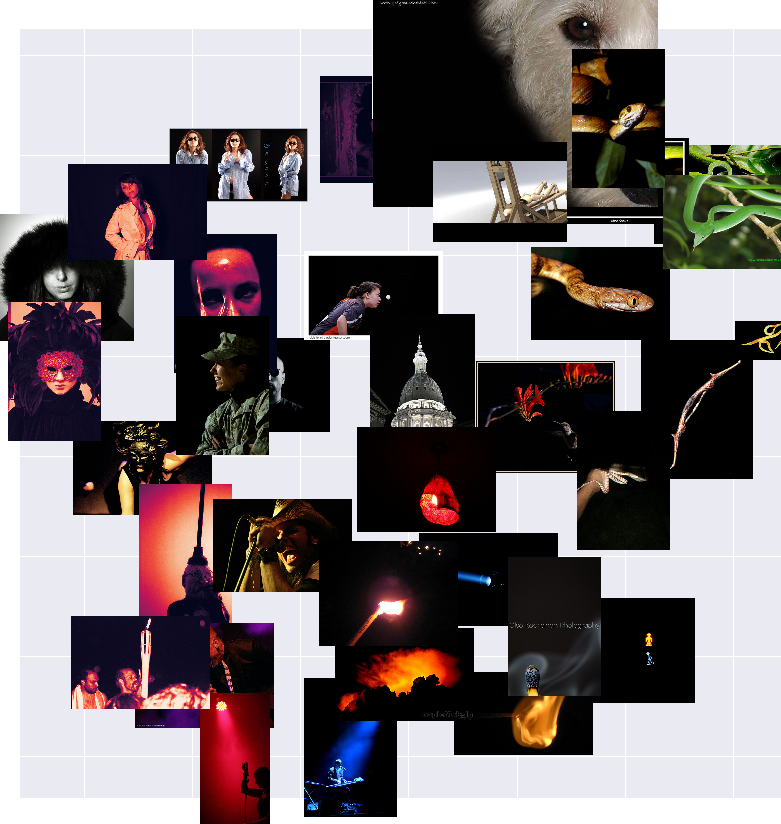}
  \caption{}
  \label{fig:kmeans_clean_13}
  \end{subfigure}
  \begin{subfigure}{0.5\linewidth}
    \includegraphics[width=1\textwidth]{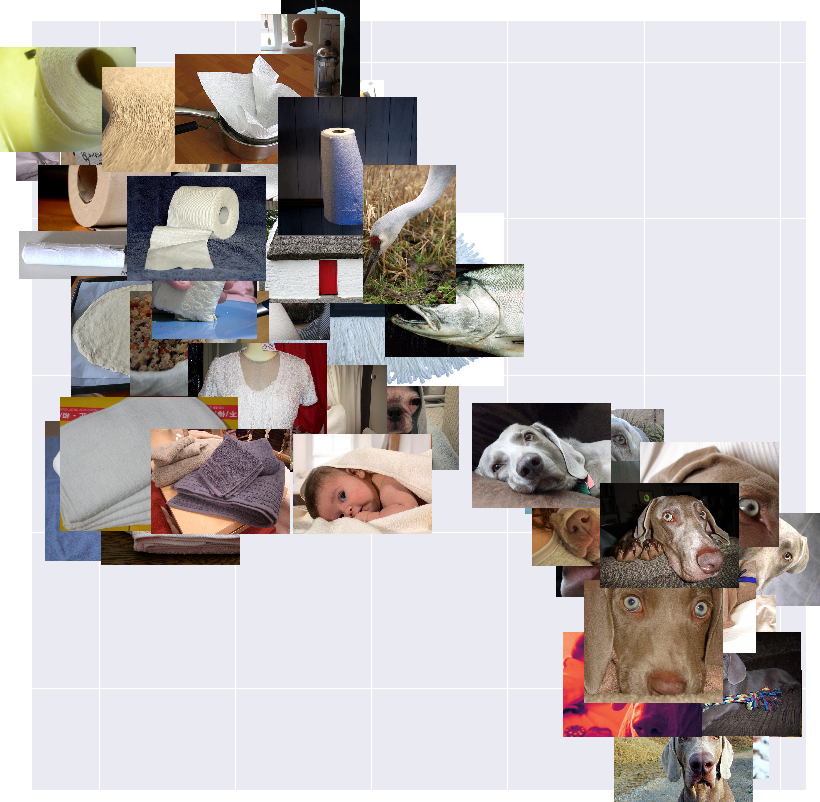}
  \caption{}
  \label{fig:kmeans_clean_27}
  \end{subfigure}
  \caption{UMAP visualisation of embeddings in latent space corresponding to images kept after k-means clusters and outlier removal for (a) polysemantic neuron 5, (b) monosemantic neuron 13, (c) neuron 1 and (d) neuron 27. }
  \label{fig:kmeans_clean_apn}
\end{figure}

\cref{fig:dendrogram_1} depicts the agglomerative clustering dendrogram for neuron 1. Neuron 1 is cut into two clusters for around 14-17 and one for $>$17. This demonstrates how we can use $d_{max}$ to control how fine-grained the concepts found are.

\begin{figure}
  \centering
    \includegraphics[width=0.5\textwidth]{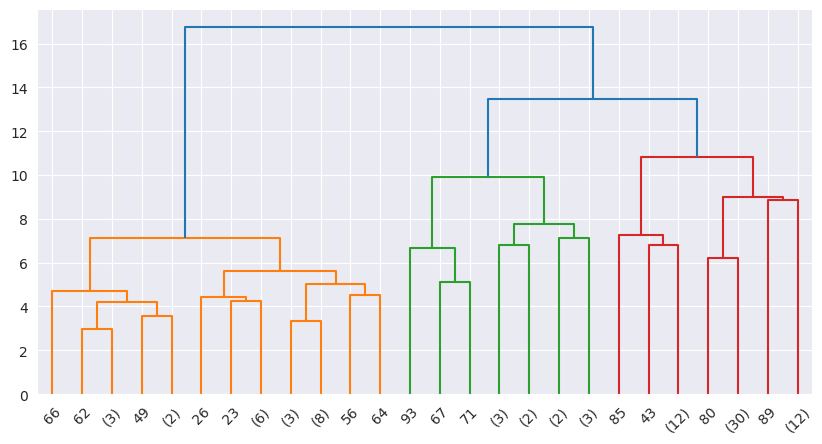}
  \caption{Neuron 1, agglomerative clustering dendrogram. }
  \label{fig:dendrogram_1}
\end{figure}

\subsection{Detail on concept vectors}
\label{apn:concept_vectors}

\cref{fig:concept_vector_35} shows the two concept vectors found for neuron 35 representing the concepts of apples and sport-type images. It can be seen from the figure that although both categories highly activate neuron 35 that some of the other activation spikes are not common in these two concepts. 

\begin{figure}
  \centering
  \begin{subfigure}{0.48\linewidth}
    \includegraphics[width=1\textwidth]{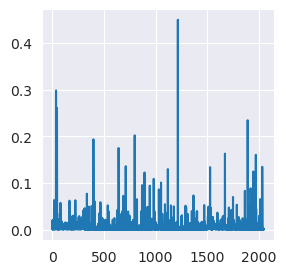}
    \caption{Concept vector 1. }
    \label{fig:concept_vector_35_cl1}
  \end{subfigure}
  \hfill
  \begin{subfigure}{0.48\linewidth}
    \includegraphics[width=1\textwidth]{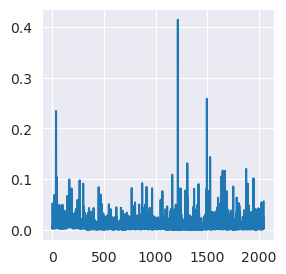}
    \caption{Concept vector 2.}
    \label{fig:concept_vector_35_cl2}
  \end{subfigure}
  \caption{Concept vectors found for neuron 35. The $x$ axis represents each dimension in activation space, so each peak is the amount that the concept vector points for the 2048 basis vectors in this particular layer. }
  \label{fig:concept_vector_35}
\end{figure}

\subsection{User Evaluation}
\label{apn:user_evaluation}

\cref{fig:user_evaluation} shows examples of questions from our user evaluation test. (a) asks if the concept conveyed by these images is described well by the concept label `curvy'. One user disagreed with this label, and instead proposed the label `sinusoidal'. 

\begin{figure}
  \centering
  \begin{subfigure}{1\linewidth}
    \includegraphics[width=1\textwidth]{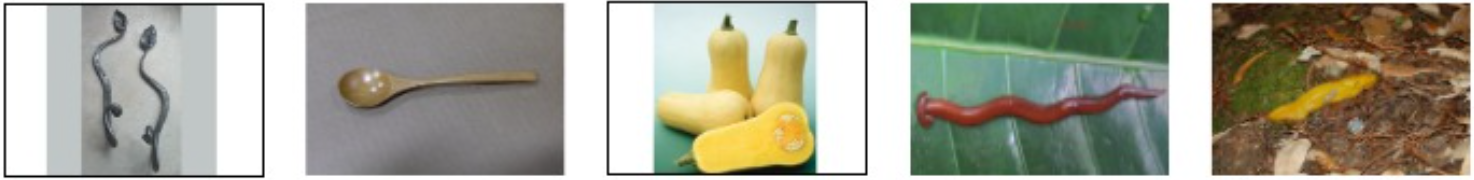}
    \caption{}
    \label{fig:curvy}
  \end{subfigure}
  \begin{subfigure}{1\linewidth}
    \includegraphics[width=1\textwidth]{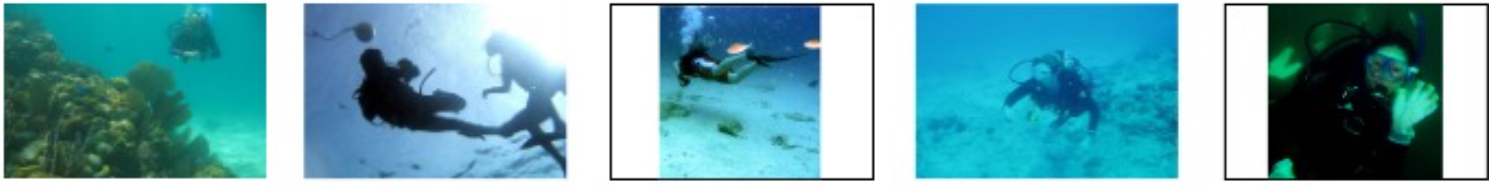}
    \caption{}
    \label{fig:diver}
  \end{subfigure}
  \hfill
  \begin{subfigure}{0.5\linewidth}
    \includegraphics[width=1\textwidth]{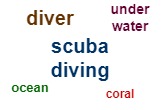}
    \caption{}
    \label{fig:wordcloud}
  \end{subfigure}
  \caption{User evaluation questions to evaluate the understandability of the semantic meaning of concepts. For (a) the user was asked if a given label  describes the concept well. Participants were asked to label the concept shown in the images in (b). The labels proposed by the users are shown in (c). The font size reflects the frequency of each label suggested by the users.}
  \label{fig:user_evaluation}
\end{figure}

}

\end{document}